# Improving Vertebra Segmentation through Joint Vertebra-Rib Atlases

Yinong Wang[a], Jianhua Yao[a], Holger R. Roth[a], Joseph E. Burns[b], and Ronald M. Summers[a]

[a]Imaging Biomarkers and Computer-Aided Diagnosis Laboratory, Radiology and Imaging Sciences, National Institutes of Health, Bethesda, MD 20892, USA;
[b]Department of Radiological Sciences, University of California-Irvine, Orange, CA 92868, USA

## ABSTRACT

Accurate spine segmentation allows for improved identification and quantitative characterization of abnormalities of the vertebra, such as vertebral fractures. However, in existing automated vertebra segmentation methods on computed tomography (CT) images, leakage into nearby bones such as ribs occurs due to the close proximity of these visibly intense structures in a 3D CT volume. To reduce this error, we propose the use of joint vertebra-rib atlases to improve the segmentation of vertebrae via multi-atlas joint label fusion. Segmentation was performed and evaluated on CTs containing 106 thoracic and lumbar vertebrae from 10 pathological and traumatic spine patients on an individual vertebra level basis. Vertebra atlases produced errors where the segmentation leaked into the ribs. The use of joint vertebra-rib atlases produced a statistically significant increase in the Dice coefficient from $92.5 \pm 3.1\%$ to $93.8 \pm 2.1\%$ for the left and right transverse processes and a decrease in the mean and max surface distance from $0.75 \pm 0.60$mm and $8.63 \pm 4.44$mm to $0.30 \pm 0.27$mm and $3.65 \pm 2.87$mm, respectively.

## 1. INTRODUCTION

The segmentation of the vertebral column is a critical step for subsequent quantitative analysis by computer aided diagnosis (CADx) systems for the purposes of detecting and treating various traumas and pathologies of the spine. Although vertebra shape is generally consistent within the spine and between individuals, traumatic injury and the onset of pathology can greatly change the morphology of structures near the spine as well as their pixel intensities on computed tomography (CT) images, making segmentation a challenging task. Although existing techniques generally perform well in segmenting vertebrae[1], segmentation often leaks into the ribs which may compromise the task of computer aided detection of fractures. For instance, false positives in a CADx system to detect posterior element fractures often occur at the costovertebral junction when ribs are mis-labeled, shown in Figure 1. Other model-based segmentation methods have been used for the task of rib cage segmentation, which includes a segmentation of the complete vertebral column with a separate label[2], suggesting that the task of rib removal is achievable. Our goal is to improve the vertebra segmentation by alleviating over-segmentation into the ribs.

## 2. METHODS

We propose a technique that utilizes a multi-atlas joint label fusion framework for the segmentation of vertebrae using spine CT images. Identification and localization of each vertebra were generated via an automated algorithm[3], which uses directed graph search, adaptive thresholding, fuzzy connectivity, and anatomic vertebral models to generate a sub-volume for each vertebra. Manual ground truth segmentations were obtained by a trained radiologist and were used to create vertebra atlases. Figure 2b shows an example of a ground truth segmentation. Rib segmentations were further included in the atlases, shown in Figure 2d, using the manual segmentation tool within the Medical Imaging Interaction Toolkit (MITK)[*].

---

[*] http://mitk.org/wiki/MITK.

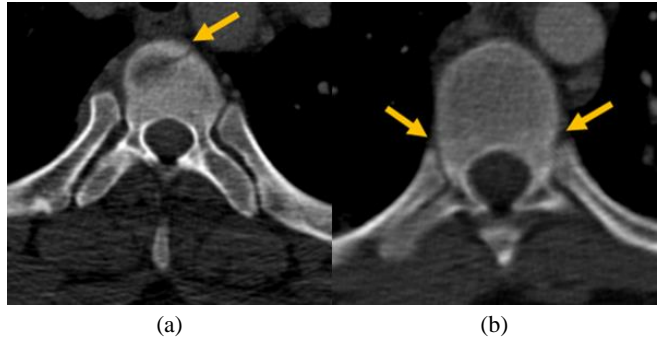

Figure 1: Axial view of (a) vertebral body fracture and (b) vertebra-rib joints mistaken as fracture lines (designated by yellow arrows).

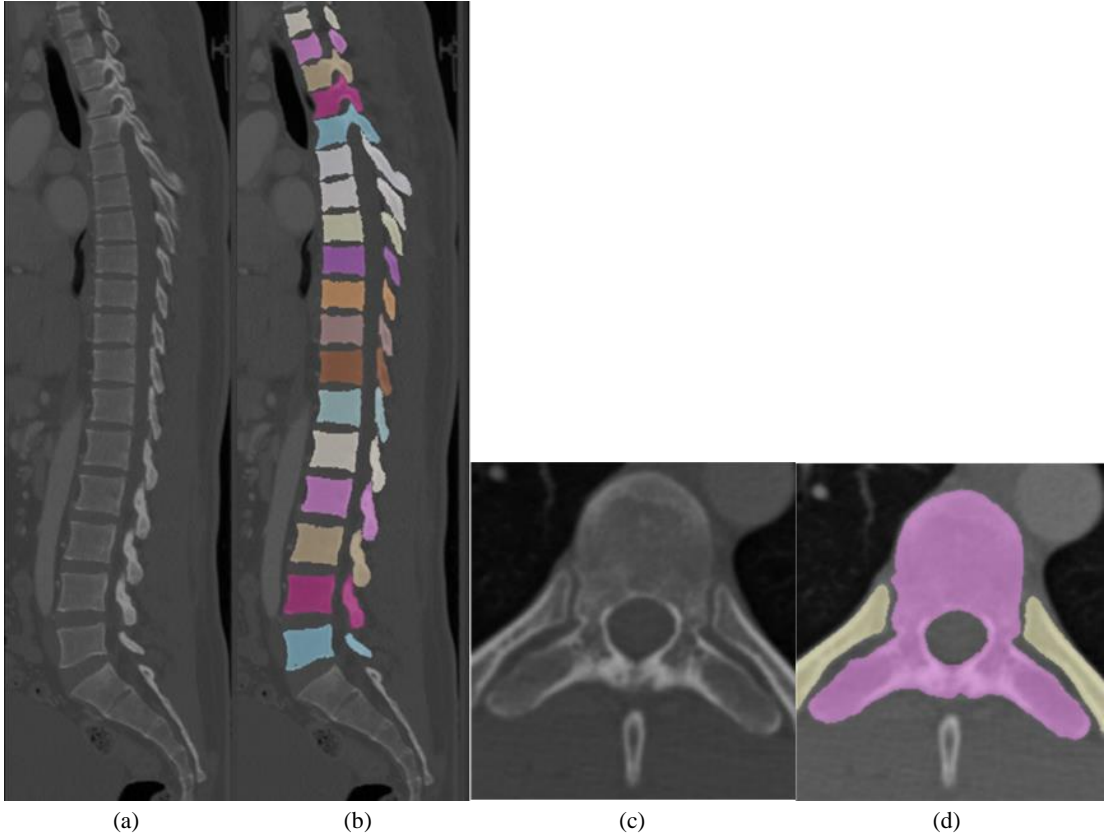

Figure 2: Sagittal view of (a) healthy spine atlas, (b) ground truth segmentation of thoracic and lumbar vertebrae, axial view of (c) vertebrae-rib atlas, and (d) ground truth segmentation.

In the multi-atlas joint label fusion framework, the atlases are first registered to the target image, initialized by affine transformation and followed by a non-rigid cubic B-spline deformable registration[4]. The symmetric affine transformation was obtained by computing an affine transformation matrix A that maximizes the intensity similarity between the target image $I_1$ and floating atlas image $I_2$ and applies the resulting transform $A$ to $I_2$ where $I_1 \cong I_2 \circ A$. Non-rigid B-spline transformation $T$ uses normalized mutual information $NMI$ between the marginal entropies $H$ of $I_1$ and $I_2 \circ T$ and joint entropy $H(I_1, \ I_2 \circ T)$ according to the following equation:

$$NMI = \frac{H(I_1) \ + \ H(I_2 \circ T)}{H(I_1, \ I_2 \circ T)} \tag{1}$$

$NMI$ was maximized by optimizing the cost function $C$ both globally and locally according to:

$$C = (1 - \alpha) * NMI - \alpha * P \tag{2}$$

Weight factor $\alpha$ was set to the default value of 0.005, and controls the bending energy-based penalty term $P$, which penalizes the non-rigid registration $T$ in order to achieve physically realistic smooth deformations. A CUDA-based GPU-accelerated open-source software package, NiftyReg[†] was used to parallelize and accelerate the registration process due to the computationally-heavy nature of the registration technique chosen. Registration results from multiple atlases were then combined via joint label fusion which uses a measure of local appearance similarity to generate a consensus binary label result[5]. Joint label fusion produced a consensus label result $S(x)$ by minimizing the expectation of combined label differences according to $\bar{S}(x)$ where $S_i(x)$ and $w_i(x)$ refer to the individual segmentations and voting weights, respectively.

$$\bar{S}(x) = \sum_{i=1}^{n} w_i(x) S_i(x) \tag{3}$$

$$w_x = \frac{M_x^{-1} * 1_n}{1_n^t M_x^{-1} 1_n} \tag{4}$$

Individual weights $w_x$ were computed using $1_n = [1; 1; ...; 1]$ and the pairwise dependency matrix $M_x$, which calculates an estimated likelihood of incorrect labels produced by two registration results on a per-voxel basis. A final morphological correction step was applied to refine the boundaries of the final label result[1]. Isolated islands of binary labels were first removed and holes were closed by morphological operators and connected components. Furthermore, collision detection and correction were necessary due to the per-vertebra nature of the registrations. A perceptron linear classifier based on voxel intensity difference and the relative distances to the centroids of the vertebrae was employed to correct instances where one voxel was assigned to segmentations of two different vertebrae. The final result then was refined using a Laplacian level set algorithm[7]. Two enhancements were implemented for the vertebra atlases. First, three vertebrae (the target and two adjacent vertebrae) were grouped to form a bundled atlas to reduce the likelihood of segmentation collision among adjacent vertebrae. Second, joint vertebra-rib atlases were created to alleviate the leakage of the segmentation result into the ribs. Examples of atlases used for registration are shown in Figure 3.

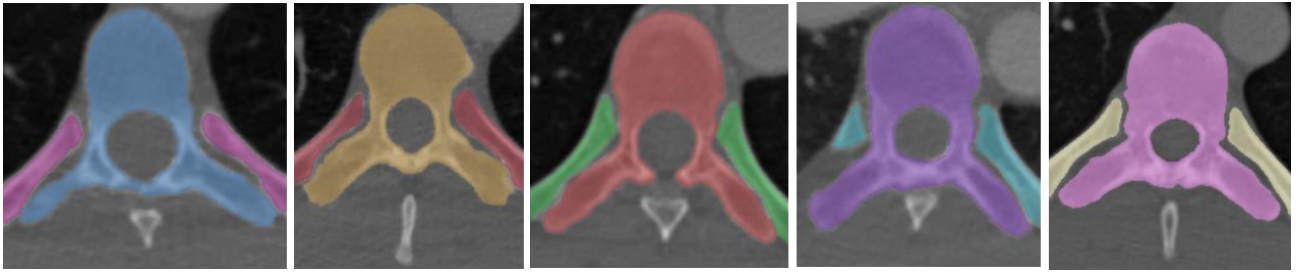

Figure 3: Axial views of joint vertebra-rib atlases (level T6).

Segmentation performance was evaluated using the Dice coefficient ($DC$) and the average and max surface distance ($ASD$, $ASD_{max}$). Dice coefficient computes a measure of similarity between the segmentation result $S$ and ground truth segmentation $T$[7]. Surface distance computes the distance between the nearest surface voxels on the segmented vertebra surface and the ground truth surface.

$$DC = \frac{2 * |GT \cap S|}{|GT| + |S|} * 100\% \qquad (5)$$

$$ASD = \frac{1}{|V_S|} \sum_{i=1}^{|V_S|} \|d_i(V_S, V_{GT})\| \qquad (6)$$

$$ASD_{max} = max(\|d_i(V_S, V_{GT})\|) \qquad (7)$$

Segmentation performance evaluation was conducted on both the whole vertebra and its substructures (vertebral body, left and right transverse processes, and spinous process) via an automated vertebra partitioning algorithm[8].

## 3. RESULTS

We evaluated our methods on patients from both pathological cases that exhibit osteopenia or osteoporosis, as well as traumatic cases where fractures of the vertebral bodies and posterior elements were identified. Five atlases were created from a cohort of healthy individuals. Two sets of atlases were generated, one with vertebra alone, the other with joint vertebra-rib atlases. Manual ground truth segmentations for 106 vertebrae on CT were conducted for performance evaluation on 10 patients, half of which were diagnosed with osteopenia or osteoporosis and the other half with traumatic fractures of the posterior elements of the vertebrae. Vertebrae from the cohort with posterior element fractures were randomly selected from instances where the vertebra segmentation leaked into the ribs using the vertebra atlases alone.

Table 1: Dice coefficient evaluation (DC): Comparison of segmentation performance using vertebra atlases (V) alone and joint vertebra-rib atlases (VR) on the whole vertebra (WV), vertebral body (VB), left and right transverse processes (TP), and spinous process (SP) from the cohorts containing posterior element fractures (PEF) and osteopenia or osteoporosis (O). Mean Dice coefficient with standard deviation (in parenthesis) reported as %. *Indicates statistical significance using paired two-tailed t-test with p-value < 0.05.

| | Number of Vertebra | DC-WV | DC-VB | DC-TP | DC-SP |
|---|---|---|---|---|---|
| **PEF-V** | 21 | 94.6 (1.5) | 95.8 (1.6) | 92.5 (3.1) | 95.1 (1.8) |
| **PEF-VR** | | 94.8 (1.6) | 95.5 (1.8) | 93.8 (2.1)* | 94.9 (1.6) |
| **O-V** | 85 | 90.8 (10.6) | 91.0 (10.3) | 90.2 (11.6) | 92.4 (10.6) |
| **O-VR** | | 90.9 (10.5) | 91.0 (10.3) | 90.3 (11.5) | 92.4 (10.6) |

A total of 21 posterior element fracture (PEF) vertebrae and 85 osteopenic and osteoporotic (O) vertebrae were examined. In instances where the vertebra segmentation leaked into the ribs, there was a statistically significant improvement in the performance using the joint vertebra-rib atlases for the left and right transverse processes (92.5% to 93.8%, p-value = 0.018), shown in Table 1. Table 2 shows statistically significant reductions for both mean and max surface distances for both the whole vertebra (0.53mm and 11.17mm to 0.33mm and 6.21mm, respectively; p-value < 0.001) and for the transverse processes (0.76mm and 8.63mm to 0.30mm and 3.65mm, respectively; p-value < 0.001). Since these improvements are more localized to the sub-regions of the rib pertaining to the transverse processes as seen visually in Figure 4, there is no statistically significant increase in Dice coefficient for the whole vertebra (94.6% to 94.8%). In addition, we observed no statistically significant change in performance in instances where there was no mislabeling of the ribs as vertebrae (90.8% to 90.9%). Figure 4c shows an example where the use of the joint vertebra-rib atlases resulted in a Dice coefficient decrease for the left transverse process, but still showed improvement on the right transverse process. The best performing result is shown in Figure 4a where the joint vertebra-rib atlases greatly improve performance on the left and right transverse processes.

Table 2: Mean (ASD) and max (ASD$_{max}$ surface distance evaluation: Comparison of segmentation performance using vertebra atlases (V) alone and joint vertebra-rib atlases (VR) on the whole vertebra (WV), vertebral body (VB), left and right transverse processes (TP), and spinous process (SP) from the cohort containing posterior element fractures (PEF). Mean and max surface distances with standard deviation (in parenthesis) reported in mm. *Indicates statistical significance using paired two-tailed t-test with p-value < 0.001.

|         | ASD -WV | ASD -VB | ASD -TP | ASD -SP | ASD$_{max}$ -WV | ASD$_{max}$ -VB | ASD$_{max}$ -TP | ASD$_{max}$ -SP |
|---------|---------|---------|---------|---------|-----------------|-----------------|-----------------|-----------------|
| PEF-V   | 0.53 (0.23) | 0.32 (0.16) | 0.76 (0.60) | 0.25 (0.31) | 11.17 (2.89) | 4.02 (1.69) | 8.63 (4.44) | 3.00 (3.11) |
| PEF-VR  | 0.33 (0.15)* | 0.36 (0.16) | 0.30 (0.27)* | 0.21 (0.17) | 6.21 (2.76)* | 4.11 (1.69) | 3.65 (2.87)* | 2.55 (2.49) |

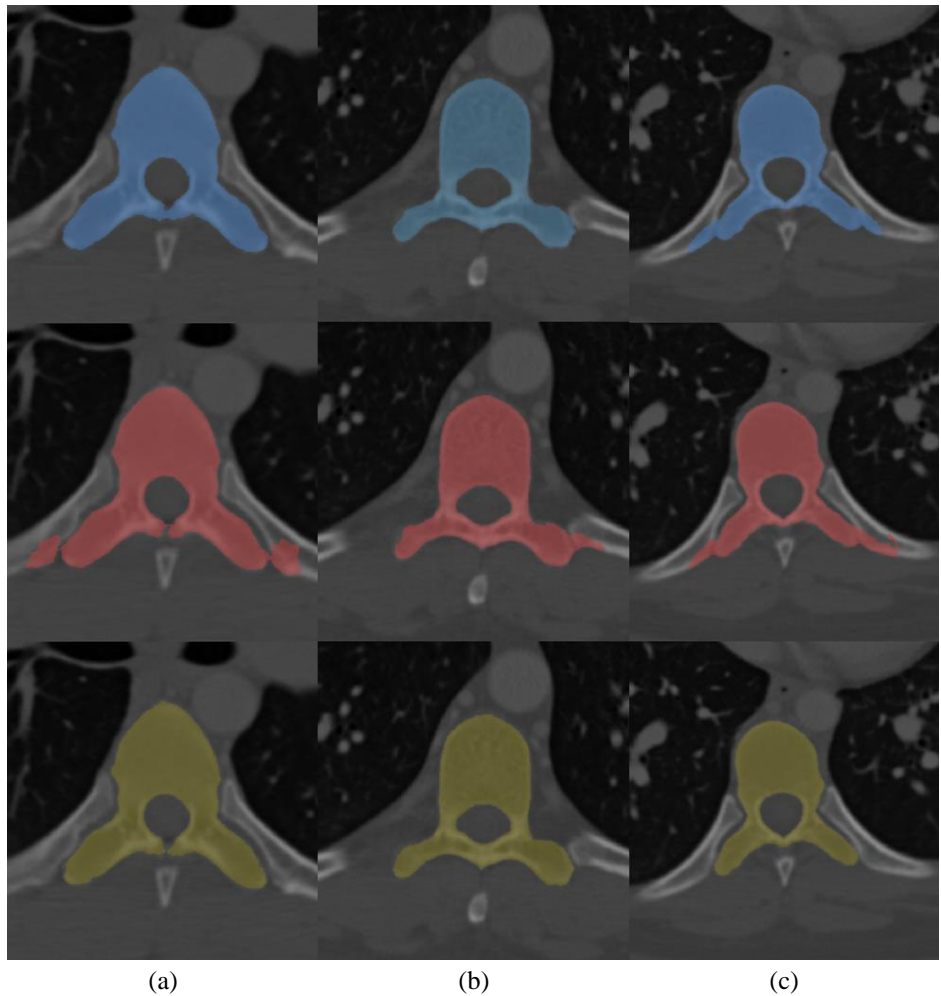

(a)   (b)   (c)

Figure 4: Ground truth segmentations (yellow) and segmentation results using vertebra atlases (red) and vertebra-rib atlases (blue). (a) Best performing (DC 89.1%, ASD 1.42mm, and ASD-max 12.57mm and 95.0%, 0.20mm, and 2.01mm for vertebra atlases and joint vertebra-rib atlases respectively), (b) average (92.9%, 0.65mm, 9.83mm and 94.2%, 0.21mm, 2.35mm for vertebra atlases and joint vertebra-rib atlases respectively), and (c) less optimal (89.8%, 1.65mm, 13.33mm and 91.0%, 0.94mm, 10.33mm for vertebra atlases and joint vertebra-rib atlases respectively) segmentation results.

## 4. DISCUSSION

The proposed method of combining multi-atlas segmentation and joint label fusion with the inclusion of rib segmentations in the spine atlases provides a significantly improved result in segmenting fractured vertebrae. In addition, the joint vertebra-rib atlases perform equivalently on osteoporotic vertebrae where there is no over-segmentation. In the set of cases with segmentation leakage into the ribs, the max surface distance improves on average by 4.98mm, which translates to a reduction in over-segmentation by a distance of nearly 16 pixels in the transverse plane. However, the method is unable to completely eliminate leakage into the ribs in 4 of the 21 vertebrae, some of which are shown in Figure 4b and 4c. This is likely due to the extremely close proximity of the edge of the transverse processes to the ribs, generally within a distance of 2 pixels.

Our results also demonstrate that the quantitative measurement of Dice coefficient alone cannot accurately represent the quality of a segmentation. In cases where the use of the joint vertebra-rib atlases dramatically improve the result both in visual inspection and in significantly decreasing the mean and maximum surface distances, the Dice coefficient often improves within a single percentage point. In other words, the Dice coefficient is more descriptive of the volume of the segmentation rather than the morphological details. For the task of eliminating ribs from the segmentation, improvements in the morphological details are more critical.

## 5. CONCLUSION

This work demonstrates that multi-atlas joint label fusion segmentation methods for vertebrae on CT benefit from using atlases containing multiple high-intensity, well-defined objects in the form of joint vertebra-rib atlases. The technique proposed in this paper reduces errors in the vertebra segmentation leaking into ribs and can potentially improve CAD performance for fracture detection in the posterior elements.

## ACKNOWLEDGEMENTS

This research was supported in part by the Intramural Research Program of National Institutes of Health, Clinical Center.